\documentclass[10pt,twocolumn,letterpaper]{article}

\usepackage{wacv}

\usepackage{array}
\usepackage{siunitx}
\usepackage{pifont}
\usepackage{multirow}
\usepackage{algorithm}
\usepackage{algpseudocode}
\usepackage{tabularx}
\usepackage[protrusion=true,expansion=true]{microtype}
\usepackage{textcomp}
\usepackage[table]{xcolor}

\newcommand{\cmark}{\ding{51}}
\definecolor{wacvblue}{rgb}{0.21,0.49,0.74}
\usepackage[pagebackref,breaklinks,colorlinks,allcolors=wacvblue]{hyperref}

\def\wacvPaperID{1239}
\def\confName{WACV}
\def\confYear{2027}

\title{
\textcolor[HTML]{0050A4}{C}%
\textcolor[HTML]{1A63B0}{o}%
\textcolor[HTML]{3376BC}{l}%
\textcolor[HTML]{4D89C8}{l}%
\textcolor[HTML]{669CD4}{a}%
\textcolor[HTML]{80AFE0}{b}%
\textcolor[HTML]{0050A4}{O}%
\textcolor[HTML]{1A63B0}{D}: Collaborative Multi-Backbone with Cross-scale Vision for UAV Small Object Detection
}

\author{
Xuecheng Bai$^{1,*,\dagger}$\begingroup

\thanks{
Equal contribution. \\
$^{\dagger}$Corresponding authors: Xuecheng Bai 
({\tt\small bai\_xuecheng@163.com}) and Chuanzhi Xu 
({\tt\small chuanzhi.xu@sydney.edu.au}).
}
\endgroup
\quad
Yuxiang Wang$^{2,*}$ \quad
Chuanzhi Xu$^{2,\dagger}$ \quad
Kang Han$^{1,3}$ \quad
Jun Guo$^{1}$ \quad
Pengfei Ye$^{4}$ \\
\\
$^{1}$Aviation Traffic Control Technology (SHENZHEN) Co., Ltd., Shenzhen, China \\
$^{2}$The University of Sydney, NSW, Australia \\
$^{3}$Research Institute of Traffic Control Technology Co., Ltd., Beijing, China \\
$^{4}$Massachusetts Institute of Technology, Cambridge, MA, USA
}

\begin{document}
\maketitle

%%%%%%%%%%%%%%%%%%%%%%%%%%%%%%%%%%%%%%%%%%%%%%%%%%%%%%%%%%%%%%%%%%%%%%%%%%%%%%%%
\begin{abstract}
Small object detection in unmanned aerial vehicle (UAV) imagery is challenging because high-altitude viewpoints produce severe scale variation, weak structural cues, and tight computational budgets. Existing lightweight detectors usually fuse multi-scale features after downsampling, where boundary and texture details have already been attenuated and heterogeneous feature streams may be spatially misaligned. To address these issues, we propose CollabOD, a collaborative detection framework that preserves structural details, aligns cross-path features before fusion, and keeps the detection head lightweight at inference time. CollabOD combines a Dual-Path Fusion Stem, a Dense Aggregation Block, a Bilateral Reweighting Module, and a Unified Detail-Aware Head to strengthen localization-oriented representation while limiting extra computation. On VisDrone, CollabOD obtains 52.4 AP$_{50}$, 30.8 AP$_{75}$, and 29.9 AP$_{50:95}$ with 65.5 GFLOPs; on UAVDT it reaches 31.2 AP$_{50}$ and 17.4 AP$_{50:95}$; and on AI-TOD it reaches 45.4 AP$_{50}$ and 20.0 AP$_{50:95}$ at 137 FPS. The code are available at: \textcolor{blue}{https://github.com/Bai-Xuecheng/CollabOD}.
\end{abstract}

%%%%%%%%%%%%%%%%%%%%%%%%%%%%%%%%%%%%%%%%%%%%%%%%%%%%%%%%%%%%%%%%%%%%%%%%%%%%%%%%
\section{INTRODUCTION}

Unmanned aerial vehicles (UAVs) have become essential platforms for autonomous perception in applications such as urban traffic monitoring~\cite{li2025ad}, traffic flow analysis~\cite{NIKOUEI2025200561}, and parking management~\cite{drones9060429}. In these scenarios, object detection plays a critical role in reliable traffic state assessment by accurately identifying and localizing targets in aerial imagery. However, high-altitude operation introduces significant scale variation, a large number of distant small objects, and limited onboard computational resources, making lightweight and accurate detection models highly desirable.

From a feature representation standpoint, small aerial objects that are typically smaller than $32 \times 32$ pixels contain extremely limited discriminative information. Their features are rapidly degraded through repeated downsampling, leading to weak representations~\cite{Luo2026UAV} and low signal-to-noise ratios~\cite{NIKOUEI2025200561}. This degradation becomes more severe under challenging aerial conditions such as low contrast~\cite{xia2026efsidetrefficientfrequencysemanticintegration}, motion blur~\cite{wang2025efficientfeaturefusionuav}, and atmospheric distortion~\cite{Luo2026UAV}. In such cases, fine-grained structural cues, e.g., object boundaries and edge textures, become critical for distinguishing foreground from background and supporting precise localization~\cite{drones9060429,Li2026MST}. Although feature pyramid networks preserve multi-scale representations, their cross-scale fusion is usually implemented via simple addition or concatenation and lacks explicit modeling of structural detail attenuation and cross-layer misalignment.

\begin{figure}[t]
  \centering
  \includegraphics[
    width=\linewidth]{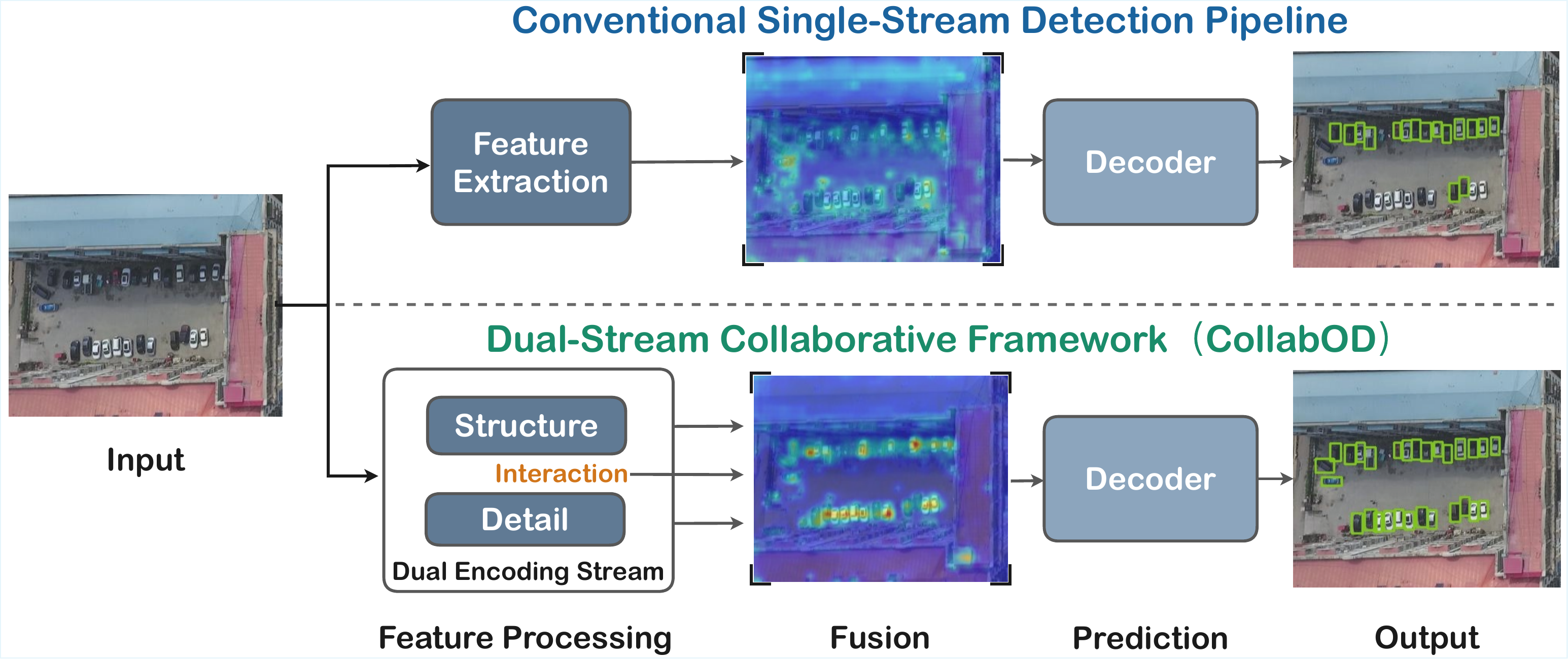}
  \caption{Comparison of conventional single-stream detection and CollabOD. Single-stream methods attenuate structural cues and perform implicit fusion, resulting in spatial misalignment. CollabOD decouples and aligns structural and detail representations prior to fusion for improved stability and accuracy.}
  \label{fig:Intro}
\end{figure}

Existing methods enhance representation capacity by introducing auxiliary branches~\cite{drones9060429}, attention mechanisms~\cite{drones9060429}, or refined fusion strategies~\cite{wang2025efficientfeaturefusionuav}. While effective, these designs often produce heterogeneous feature streams with distinct receptive field distributions and semantic biases. Conventional fusion implicitly assumes spatial and semantic compatibility across paths, making it difficult to explicitly suppress cross-path discrepancies. For small objects with inherently limited structural and semantic representations, even minor spatial misalignment can be amplified during bounding box regression, leading to localization instability and degraded robustness. Consequently, UAV small object detection is jointly constrained by structural detail attenuation and implicit cross-path fusion inconsistency.

To address these issues, we enhance structural detail preservation and calibrate heterogeneous feature streams prior to fusion, as shown in Fig.~\ref{fig:Intro}. First, localization-related structural cues should be strengthened under lightweight constraints. Second, heterogeneous feature streams should be calibrated before fusion to improve spatial and semantic compatibility.

\begin{figure*}[t]
  \centering
  \includegraphics[
    width=\textwidth]{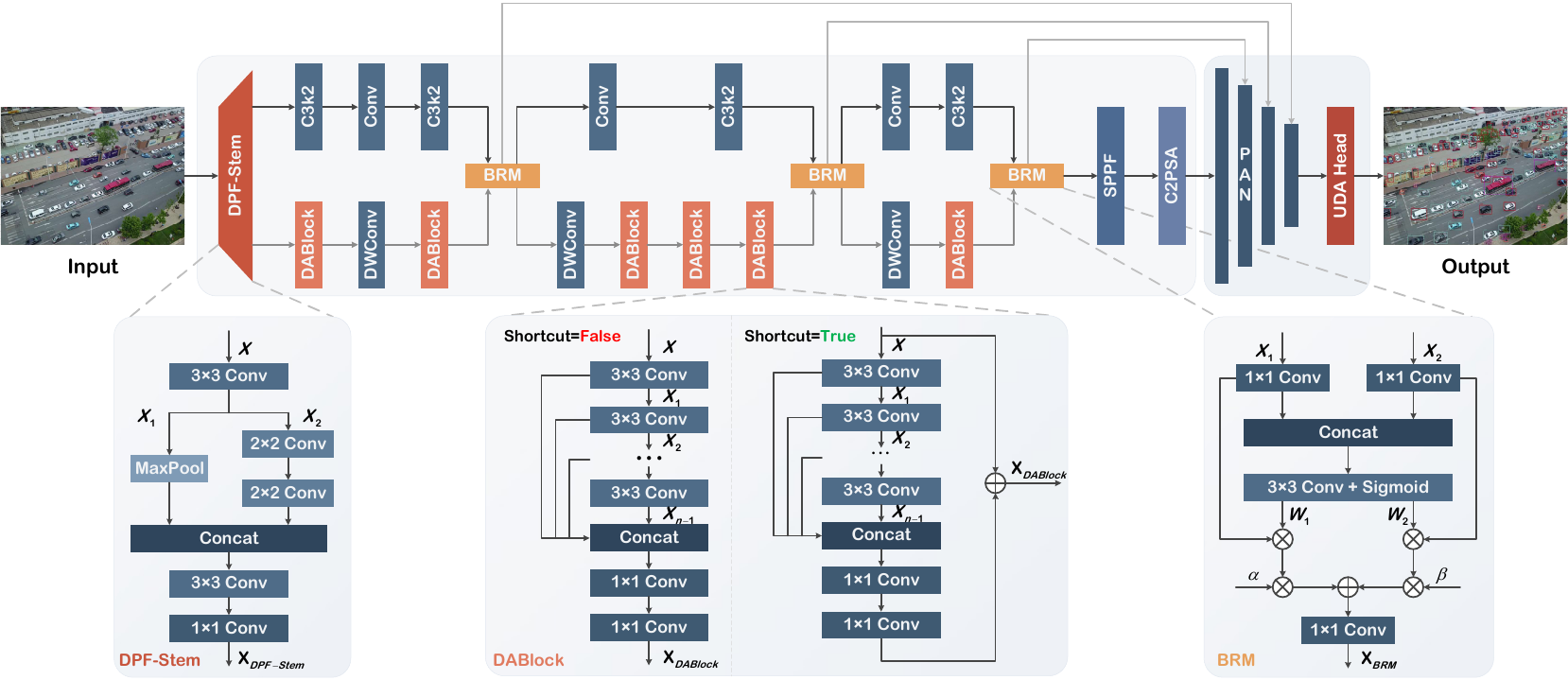}
  \caption{Overview of the proposed CollabOD framework. DPF-Stem denotes the Dual-Path Fusion Stem, DABlock denotes the Dense Aggregation Block, and BRM denotes the Bilateral Reweighting Module. The UDA Head corresponds to the Unified Detail-Aware Head, which is detailed in Section~\ref{sec:localization_aware_lightweight_design}. The remaining components are inherited from the original YOLO11 architecture.}
  \label{fig:framework}
\end{figure*}

In this paper, we propose \textbf{\underline{CollabOD}}, a \underline{collab}orative small \underline{o}bject \underline{d}etection framework built upon YOLO11-M-P2~\cite{yolo11_ultralytics}. Fig.~\ref{fig:framework} summarizes the overall architecture of CollabOD. CollabOD systematically improves input encoding, backbone representation, multi-scale fusion, and detection head design to achieve structural detail enhancement, cross-path alignment, and computational efficiency. Experiments on VisDrone~\cite{vsdrone}, UAVDT~\cite{uavdt}, and AI-TOD~\cite{wang2021tiny} show that CollabOD improves detection robustness in challenging aerial scenes. Within the reported comparisons, it attains the highest AP\textsubscript{75} on VisDrone while using the lowest GFLOPs among methods with complete efficiency reports, obtains the best AP\textsubscript{50} and AP\textsubscript{50:95} on UAVDT, and achieves the best AP\textsubscript{50}, AP\textsubscript{50:95}, GFLOPs, and FPS among the evaluated YOLO-series baselines on AI-TOD.

The contributions of this work are summarized as follows:

\begin{itemize}[itemsep=2pt, topsep=0pt, parsep=0pt]
    \item We develop a lightweight detection framework \textbf{CollabOD} that jointly enhances structural details and aligns heterogeneous feature streams, ensuring stable localization and high detection accuracy for small objects under limited computational budgets.
    
    \item We design a \textbf{Dual-Path Fusion Stem (DPF-Stem)} and a \textbf{Dense Aggregation Block (DABlock)} to mitigate the progressive degradation of localization-related structural information in deep networks, preserving boundary and contour cues at the input stage while compensating for hierarchical feature attenuation.
    
    \item We introduce a \textbf{Bilateral Reweighting Module (BRM)} that improves cross-path feature consistency through channel-wise adaptive weight generation and learnable scaling.
    
    \item We propose a \textbf{Unified Detail-Aware Head (UDA Head)} that enhances boundary regression via detail-aware convolution and employs re-parameterization to eliminate additional inference overhead.
\end{itemize}

\section{RELATED WORK}
This section reviews recent advances in UAV small object detection, focusing on structural representation, cross-scale feature learning, and efficiency-aware localization design from the perspective of localization stability under deployment constraints.
\subsection{Structural Representation for Small Objects}
For small object detection in UAV aerial imagery, structural representation capability can be characterized along two dimensions: the supply strength of structural information for small objects, and the stability of structural features during hierarchical propagation. Related research has primarily evolved along two corresponding paths: detail supply and structural compensation.

At the information supply level, early methods typically increase effective pixels through higher input resolution~\cite{liu2024esod}, slice- or patch-based inference~\cite{obss2021sahi}, or super-resolution assistance~\cite{zhang2023superyolo}, but offer limited mitigation of structural information degradation caused by deep downsampling. More recent detection frameworks explicitly introduce higher-resolution feature layers~\cite{li2025ad} or adjust pyramid allocation strategies~\cite{yang2023afpn} to preserve fine-grained structures, thereby enhancing the initial representational capacity of structural information.

At the propagation stability level, advanced techniques focus on edge-sensitive enhancement~\cite{lu2025legnet}, local context modeling~\cite{wang2025mgdfis}, and multi-path representation design~\cite{chen2025yolo} to strengthen the cross-layer transmission of structural cues, shifting small object representation from single-path enhancement toward multi-source collaborative expression. The prevailing trend suggests that jointly improving structural information supply and propagation stability within a unified framework can more robustly support fine-grained localization of small objects in aerial imagery.

However, under lightweight deployment constraints in UAV systems, how to explicitly enhance localization-related structural information remains a key issue.

\subsection{Cross-Scale and Multi-Branch Feature Learning}
The core objective of cross-scale and multi-branch designs is to enhance feature interaction and improve the representational stability of small objects in complex scenes. Early methods, represented by FPN~\cite{lin2017feature}, achieve progressive fusion of multi-scale features through hierarchical pyramid structures. PANet~\cite{wang2019panet} and PAFPN~\cite{liu2018path} further strengthen bidirectional information flow, while NAS-FPN~\cite{ghiasi2019fpn} and ASF~\cite{liu2019learning} improve cross-scale integration flexibility through adaptive redistribution and structural optimization.

With the evolution of network architectures, multi-branch detection frameworks and multi-backbone designs~\cite{yang2025mhaf} introduce parallel representation pathways that enhance feature diversity through path-specific structures and explicit interaction mechanisms. MoE~\cite{lin2025yolo}, cross-branch gating~\cite{11082355}, and collaborative distillation models~\cite{yang2022focal} further model inter-path information selection and synergy, moving feature fusion from implicit aggregation toward explicit interaction and dynamic collaboration.

Despite the continuous evolution of cross-scale feature interaction mechanisms, existing methods still provide limited modeling of consistency between heterogeneous feature streams prior to fusion, which in UAV scenarios may amplify spatial and semantic misalignment and thus impair fine-grained localization stability.

\subsection{Localization and Efficient Detection Design}
Feature representation and interaction mechanisms must ultimately translate into stable localization and efficient inference. To improve regression quality, modern detectors adopt decoupled classification and regression branches~\cite{zhuang2023task} and integrate IoU-based losses such as GIoU~\cite{rezatofighi2019generalized}, EIoU~\cite{zhang2022focal}, and DIoU~\cite{zheng2020distance} to enhance bounding box stability. Given the sensitivity of small objects to fine-grained structural cues, several approaches further refine regression design or strengthen boundary-aware representations. Researchers also use structural re-parameterization and lightweight backbones to balance efficiency and representational capacity for computationally constrained deployment.

Taken together, a collaborative mechanism across representation, interaction, and prediction becomes increasingly important for robust deployment.

\section{METHODOLOGY}
In this section, we present CollabOD, a lightweight small object detection framework for UAV imagery, as shown in Fig.~\ref{fig:framework}.
Considering the instability introduced by structural degradation and cross-path inconsistency under lightweight deployment constraints, we focus on two aspects: enhancing localization-related structural information and improving the consistency of heterogeneous feature streams prior to fusion.
Accordingly, the proposed framework consists of three collaborative components: Structural Detail Preservation, Cross-Path Feature Alignment, and Localization-Aware Lightweight Design. These mechanisms are discussed in Sections~\ref{sec:structural_detail_preservation},~\ref{sec:cross_path_feature_alignment}, and~\ref{sec:localization_aware_lightweight_design}, respectively, with experimental validation presented in Section~\ref{sec:experiments}. We use crowded UAV traffic scenes as a running case, where tiny vehicles require stable structural cues and calibrated feature fusion for precise localization.

\subsection{Structural Detail Preservation}
\label{sec:structural_detail_preservation}
Small object cues critical for precise localization in UAV imagery mainly reside in boundary contours and texture gradients. However, repeated downsampling in deep backbones progressively attenuates such high-frequency responses. To mitigate structural decay at both the input and backbone stages, we design a \textbf{Dual-Path Fusion Stem (DPF-Stem)} for early preservation and a \textbf{Dense Aggregation Block (DABlock)} for hierarchical compensation.

\paragraph{Dual-Path Fusion Stem}
Given the input feature $X \in \mathbb{R}^{C \times H \times W}$, the core principle of the DPF-Stem is to partition the features into two complementary streams: a structure stream and a detail stream. First, the input feature is embedded and split as Eq.~\eqref{eq:dpf_split}:
\begin{equation}
\label{eq:dpf_split}
\{X_s, X_d\} = \mathrm{Split}(\phi(X)),
\end{equation}
where $\phi(\cdot)$ denotes a lightweight feature embedding, and $\mathrm{Split}(\cdot)$ represents a channel-wise splitting operator. These two streams are respectively responsible for preserving low-frequency geometric contours and high-frequency texture gradients, as summarized in Eq.~\eqref{eq:dpf_streams}:
\begin{equation}
\label{eq:dpf_streams}
Z_s = \Psi_{\text{pool}}(X_s), \quad Z_d = \Psi_{\text{conv}}(X_d),
\end{equation}
where $\Psi_{\text{pool}}$ employs max projection or pooling to aggregate stable structural responses, and $\Psi_{\text{conv}}$ is a learnable lightweight convolution designed to preserve texture gradients and local differential responses. In our implementation, $\phi(\cdot)$ is a Conv-BN-SiLU embedding that maps the input to $C$ channels and splits it evenly into $C/2$ structural channels and $C/2$ detail channels. The branch operators are written in Eqs.~\eqref{eq:dpf_pool_op} and~\eqref{eq:dpf_conv_op}:
\begin{align}
\Psi_{\text{pool}}(X_s) &= \mathrm{MP}_{2,2}(X_s),
\label{eq:dpf_pool_op}\\
\Psi_{\text{conv}}(X_d) &=
\mathrm{PWConv}_{1\times1}\left(\mathrm{DWConv}_{3\times3}(X_d)\right).
\label{eq:dpf_conv_op}
\end{align}
where $\mathrm{MP}_{2,2}$ denotes stride-2 max pooling, $\mathrm{DWConv}$ denotes depth-wise convolution, and $\mathrm{PWConv}$ denotes point-wise projection. This formulation separates stable geometric responses from local differential responses before the first fusion operation. The structural branch applies $2\times2$ max pooling with stride 2, while the detail branch uses a lightweight $3\times3$ depth-wise convolution with stride 2 followed by a $1\times1$ point-wise projection. The two streams are then fused at the same scale to obtain the stem output in Eq.~\eqref{eq:dpf_fusion}:
\begin{equation}
\label{eq:dpf_fusion}
X_{\text{DPF}} = \phi_{\text{fuse}}(Z_s \oplus Z_d),
\end{equation}
where $\oplus$ denotes concatenation, and $\phi_{\text{fuse}}$ is a lightweight projection used for channel mixing and scale alignment. This dual-stream modeling ensures that the DPF-Stem preserves high-frequency structural responses before and after downsampling, thereby mitigating the loss of early structural information.

\paragraph{Dense Aggregation Block}
Even though structural details are preserved at the input stage, they still suffer from gradual attenuation during repeated downsampling and cross-layer propagation within deep networks. The objective of the DABlock is to compensate for this hierarchical structural attenuation within the backbone by continuously injecting shallow, fine-grained structural responses into deeper features via dense aggregation. Let $\{X_i\}_{i=1}^n$ denote feature maps from preceding stages aligned to the current scale. The DABlock aggregates these features and refines them via stacked convolutions as Eq.~\eqref{eq:dablock_aggregation}:
\begin{equation}
\label{eq:dablock_aggregation}
X_{\text{DABlock}} =
\psi^{(2)}_{\text{conv}}\left(\bigoplus_{i=1}^{n} X_i\right)
+ \delta X,\quad \delta \in \{0,1\}.
\end{equation}

Here, $\oplus$ denotes channel concatenation, and $X$ is the current-stage input. The set $\{X_i\}_{i=1}^{n}$ contains the current feature and the aligned features from preceding dense branches at the same resolution. To make the aggregation well defined across stages, each incoming feature is first transformed by the scale-alignment operator in Eq.~\eqref{eq:dablock_alignment}:
\begin{equation}
\label{eq:dablock_alignment}
\begin{aligned}
\tilde{X}_i &= \mathcal{A}_i(X_i)\\
&= \mathrm{Conv}_{1\times1}
\left(\mathrm{Resize}(X_i; H,W)\right),
\quad i=1,\ldots,n .
\end{aligned}
\end{equation}
where $\mathrm{Resize}(\cdot;H,W)$ adapts the spatial resolution to the current stage and the $1\times1$ projection normalizes the channel dimension before concatenation. With this notation, dense aggregation is equivalently implemented as $\psi^{(2)}_{\text{conv}}(\oplus_i \tilde{X}_i)$, followed by the optional residual path $\delta X$. The operator $\psi^{(2)}_{\text{conv}}$ consists of two Conv-BN-SiLU layers: a $1\times1$ convolution for channel compression and a $3\times3$ convolution for local structural refinement. The residual switch $\delta$ preserves identity propagation when the input and output dimensions match. This design reinforces shallow structural cues in deeper representations and mitigates detail attenuation.
\begin{figure}[t]
  \centering
  \includegraphics[
    width=\linewidth]{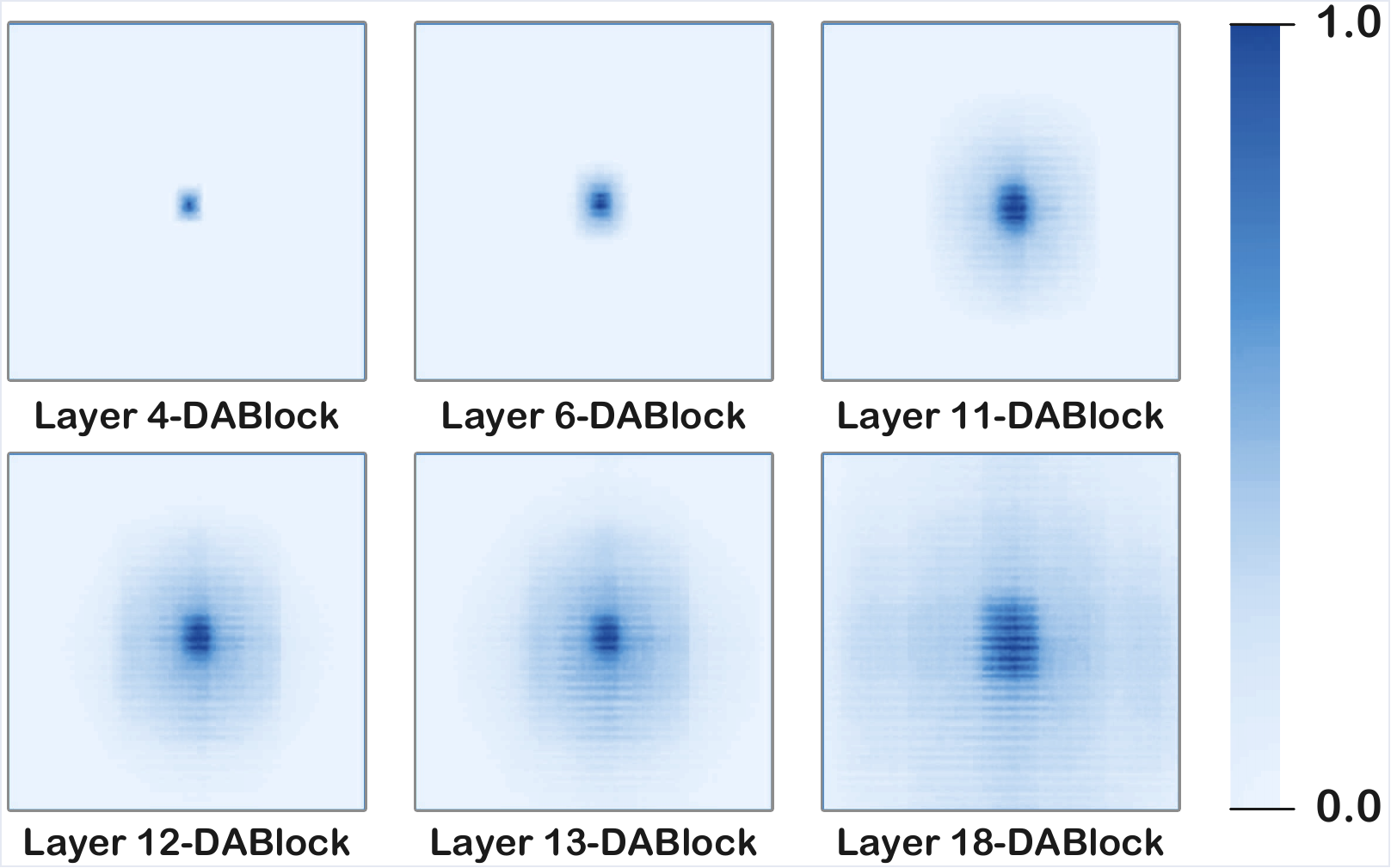}
  \caption{Effective receptive field visualization across DABlock layers. Deeper DABlock layers broaden high-contribution regions while retaining concentrated responses around object structures.}
  \label{fig:ERF}
\end{figure}

From an ERF perspective, dense aggregation promotes progressive spatial interaction across layers. By integrating aligned multi-level features, DABlock enhances long-range dependency modeling while preserving structural details. As shown in Fig.~\ref{fig:ERF}, the ERF expands steadily with depth, exhibiting increasingly broader high-contribution regions.

\subsection{Cross-Path Feature Alignment}
\label{sec:cross_path_feature_alignment}
To mitigate the inconsistency between heterogeneous feature streams before fusion, we construct the \textbf{Bilateral Reweighting Module (BRM)} to calibrate two-stream features before multiple backbone pathways are fused. Specifically, given two-stream features at the same scale $X^{(1)}, X^{(2)} \in \mathbb{R}^{C \times H \times W}$, we first map the features into a unified embedding space using a lightweight projection (e.g., $1\times1$ convolution) to obtain $\hat{X}^{(1)}$ and $\hat{X}^{(2)}$. Eq.~\eqref{eq:brm_joint_embedding} then embeds them jointly across pathways to capture the joint context:
\begin{equation}
\label{eq:brm_joint_embedding}
Z = \psi\left(\left[\hat{X}^{(1)}, \hat{X}^{(2)}\right]\right),
\end{equation}
where $[\cdot, \cdot]$ denotes concatenation along the channel dimension, and $\psi$ is a lightweight spatial interaction operator used to model cross-pathway dependencies. The bilateral gating masks are generated via activation and splitting in Eq.~\eqref{eq:brm_masks}:
\begin{equation}
\label{eq:brm_masks}
\left[G^{(1)}, G^{(2)}\right] = \mathrm{Split}\left(\sigma(Z)\right),
\end{equation}
where $\sigma$ represents the sigmoid activation function, and $\mathrm{Split}(\cdot)$ evenly divides the channels to obtain the two-stream masks $G^{(k)}\in(0,1)^{C\times H \times W}$. Unlike channel-only gating, these masks are spatially dependent, enabling finer suppression of cross-pathway redundancy and biased responses under complex backgrounds.

After obtaining the bilateral masks, the BRM reweights the two streams and achieves statistical scale calibration prior to fusion via the learnable channel amplitude modulation in Eq.~\eqref{eq:brm_fusion}:
\begin{equation}
\label{eq:brm_fusion}
\begin{aligned}
X_{\text{BRM}}
&= \phi_{\text{out}}\left(
\sum_{k=1}^{2}
\hat{X}^{(k)} \odot G^{(k)}
\odot \lambda^{(k)}
\right),
\end{aligned}
\end{equation}
where $\odot$ denotes the Hadamard product; $\lambda^{(k)}\in \mathbb{R}^{C \times 1 \times 1}$ is a learnable channel scaling factor that balances the response amplitudes of the two streams and stabilizes gradient flow; and $\phi_{\text{out}}$ is a $1\times1$ projection used for channel mixing and output integration. Through bilateral spatial reweighting and channel calibration, the BRM alleviates cross-pathway discrepancies before fusion, improving feature compatibility and stabilizing the subsequent localization regression.

The BRM can be interpreted as a pre-fusion compatibility calibration step rather than a generic attention layer. Standard channel attention estimates a single response vector from one feature stream and then rescales that stream independently. In contrast, BRM estimates two spatial masks from the joint embedding $Z$, so each path is reweighted with awareness of the other path before aggregation. For compact notation, let $\mathbf{x}^{(k)}_{u,v}$ and $\mathbf{g}^{(k)}_{u,v}$ denote the channel vectors of $\hat{X}^{(k)}$ and $G^{(k)}$ at spatial location $(u,v)$. Eqs.~\eqref{eq:brm_spatial_output} and~\eqref{eq:brm_spatial_response} give the calibrated response at this location:
\begin{align}
\mathbf{x}^{\text{BRM}}_{u,v}
&= \phi_{\text{out}}\left(\mathbf{r}_{u,v}\right),
\label{eq:brm_spatial_output}\\
\mathbf{r}_{u,v}
&=
\sum_{k=1}^{2}
\lambda^{(k)} \odot \mathbf{g}^{(k)}_{u,v}
\odot \mathbf{x}^{(k)}_{u,v}.
\label{eq:brm_spatial_response}
\end{align}
This shows that the module performs location-specific path selection and channel-wise amplitude correction simultaneously. The learnable amplitudes $\lambda^{(k)}$ further compensate for path-specific scale differences, which reduces the risk that one backbone dominates the fused representation solely because its activation magnitude is larger.

\subsection{Localization-Aware Lightweight Design}
\label{sec:localization_aware_lightweight_design}
Following structural enhancement and cross-path calibration, the remaining challenge is to enable the regression head to use these structural cues without adding inference-time overhead. To this end, the proposed \textbf{Unified Detail-Aware Head (UDA Head)} balances localization stability and efficiency through shared detail enhancement and decoupled prediction. Algorithm~\ref{alg:uda} summarizes the forward process, where multi-scale features are denoted as $F_i$.
\begin{algorithm}[t]
\caption{UDA Head: Unified Detail-Aware Head}
\label{alg:uda}
{\color{NavyBlue}\textbf{Input:}} Multi-scale features $\{F_i\}_{i\in\{xs,s,m,l\}}$, 
class number $N_c$, DFL bins $R$, hidden dimension $C_h$\\
{\color{NavyBlue}\textbf{Output:}} Decoded bounding boxes $B$ and classification scores $S$
\vspace{0.3em}
\begin{algorithmic}[1]
\State {\color{Gray}// Shared projection and detail enhancement}
\For{$i\in\{xs,s,m,l\}$}
    \State $G_i \leftarrow \mathcal{S}(\mathrm{Conv}_{1\times1}(F_i))$
    \State $P_i \leftarrow 
    \mathrm{Concat}(s_i\mathcal{H}_{box}(G_i),
    \mathcal{H}_{cls}(G_i))$
\EndFor
\vspace{0.2em}
\State {\color{Gray}// Flatten and merge multi-scale predictions}
\State $Q \leftarrow 
\mathrm{Concat}_{i\in\{xs,s,m,l\}}
(\mathrm{Reshape}(P_i))$
\vspace{0.2em}
\State {\color{Gray}// Distribution Focal Loss decoding}
\State $(B_{\text{raw}}, C_{\text{raw}}) 
\leftarrow \mathrm{Split}(Q,\{4R,N_c\})$
\State $D \leftarrow \mathrm{DFL}(B_{\text{raw}})$
{\color{Orange}$\triangleright$\ convert distributions to distances}
\vspace{0.2em}
\State {\color{Gray}// Bounding box decoding}
\State $B \leftarrow \mathrm{Dist2BBox}(D)$
\State $S \leftarrow \sigma(C_{\text{raw}})$
\vspace{0.1em}
\State \Return $\mathrm{Concat}(B,S)$
\end{algorithmic}
\end{algorithm}

% ================= Complexity Analysis =================

\paragraph{Re-parameterized Detail Enhancement}
The shared block $\mathcal{S}$ is implemented as a training-time multi-branch detail enhancer with a $3\times3$ convolution branch, a $1\times1$ convolution branch, and an identity branch when the input and output dimensions match. Each branch is followed by batch normalization during training. For an input feature $x$, this block is expressed as Eq.~\eqref{eq:uda_training_block}:
\begin{equation}
\label{eq:uda_training_block}
\begin{aligned}
\mathcal{S}(x)=
\mathrm{BN}_3(W_3*x)+
\mathrm{BN}_1(W_1*x)+
\mathrm{BN}_0(x),
\end{aligned}
\end{equation}
where $W_3$ and $W_1$ are the $3\times3$ and $1\times1$ kernels, and $\mathrm{BN}_0$ denotes the identity branch with batch normalization. Before inference, batch-normalization parameters are folded into the corresponding convolution kernels. The $1\times1$ kernel is then padded to $3\times3$, and all branch kernels and biases are summed into a single $3\times3$ convolution as Eqs.~\eqref{eq:uda_reparam_weight} and~\eqref{eq:uda_reparam_bias}:

\begin{align}
W_{\text{rep}} &=
\tilde{W}_3+\mathrm{Pad}(\tilde{W}_1)+\tilde{W}_0,
\label{eq:uda_reparam_weight}\\
b_{\text{rep}} &=
\tilde{b}_3+\tilde{b}_1+\tilde{b}_0.
\label{eq:uda_reparam_bias}
\end{align}
Thus, $\mathcal{S}(x)$ is replaced by $W_{\text{rep}}*x+b_{\text{rep}}$ at inference. As a result, the UDA Head uses the multi-branch representation during training but introduces no extra branch evaluation at deployment.

\paragraph{Complexity Analysis}
The primary computational overhead comes from the shared detail enhancement block $\mathcal{S}$ and the prediction heads. Let $N = \sum_i H_i W_i$ denote the total number of spatial locations across all scales. The time complexity can be formulated as Eq.~\eqref{eq:uda_time_complexity}:

\begin{equation}
\label{eq:uda_time_complexity}
\mathcal{O}\big(N C_h^2\big) + \mathcal{O}\big(N C_h\big) + \mathcal{O}\big(N R\big),
\end{equation}

Typically, since $C_h \gg R$, the dominant term is $\mathcal{O}\big(N C_h^2\big)$. The space complexity is primarily composed of the intermediate features and the prediction logits in Eq.~\eqref{eq:uda_space_complexity}:

\begin{equation}
\label{eq:uda_space_complexity}
\mathcal{O}\big(N C_h\big) + \mathcal{O}\big(N(4R + N_c)\big).
\end{equation}

Because $\mathcal{S}$ and the projections are shared across all scales and re-parameterized before inference, the UDA Head enhances detail perception for regression while maintaining low additional parameter counts and inference overhead. This property makes it suitable for UAV deployment scenarios with constrained computational resources.

\section{EXPERIMENTS AND RESULTS}
\label{sec:experiments}

\subsection{Experimental Settings}
\subsubsection{Implementation Details} CollabOD is built upon the YOLO11-M-P2 architecture. All experiments were conducted on an NVIDIA RTX 5090D GPU. We used the SGD optimizer with an initial learning rate of 0.01 and momentum of 0.937. The input image size was 640$\times$640, the batch size was 8, and each model was trained for 500 epochs. For the YOLO-series baselines in our controlled experiments, we used the same input size, training schedule, and evaluation code. For previously published methods, we report the numbers available in the corresponding papers and avoid drawing conclusions from metrics that were not reported under the same protocol.

\subsubsection{Dataset} We evaluate on three widely used UAV object detection benchmarks, namely VisDrone-2019-DET~\cite{vsdrone}, UAVDT~\cite{uavdt}, and AI-TOD~\cite{wang2021tiny}.
\underline{\textbf{VisDrone-2019-DET}}~\cite{vsdrone} is a widely adopted UAV detection benchmark comprising 10,209 images captured across diverse cities, times, and flight altitudes, covering 10 categories (e.g., van, truck, awning-tricycle) with significant scale variation and complex backgrounds. \underline{\textbf{UAVDT}}~\cite{uavdt} is a large-scale traffic-oriented UAV benchmark containing 77,819 annotated frames extracted from 100 video sequences, covering four vehicle categories (car, truck, bus, and other vehicle) across urban roads, intersections, and highways, with rich attribute annotations including weather, altitude, occlusion, and illumination conditions. \underline{\textbf{AI-TOD}}~\cite{wang2021tiny} is a dedicated remote sensing benchmark tailored for tiny object detection, containing 28,036 images with 700,621 annotated instances. It covers eight object categories, including bridge, ship, vehicle, storage tank, person, swimming pool, windmill, and airplane.

\subsubsection{Metrics} We adopt the standard COCO evaluation protocol~\cite{lin2014microsoft}, reporting AP$_\text{S}$, AP$_\text{M}$, AP$_{50}$, AP$_{75}$, and AP$_{50:95}$ (averaged over IoU from 0.5 to 0.95 with a step of 0.05) to evaluate detection consistency across varying localization strictness. Since AP$_{50}$ uses a less strict IoU threshold than AP$_{75}$, we verify that every reported result satisfies AP$_{50}\geq$AP$_{75}$ before analysis.

\begin{figure*}[t]
  \centering
  \includegraphics[
    width=\textwidth]{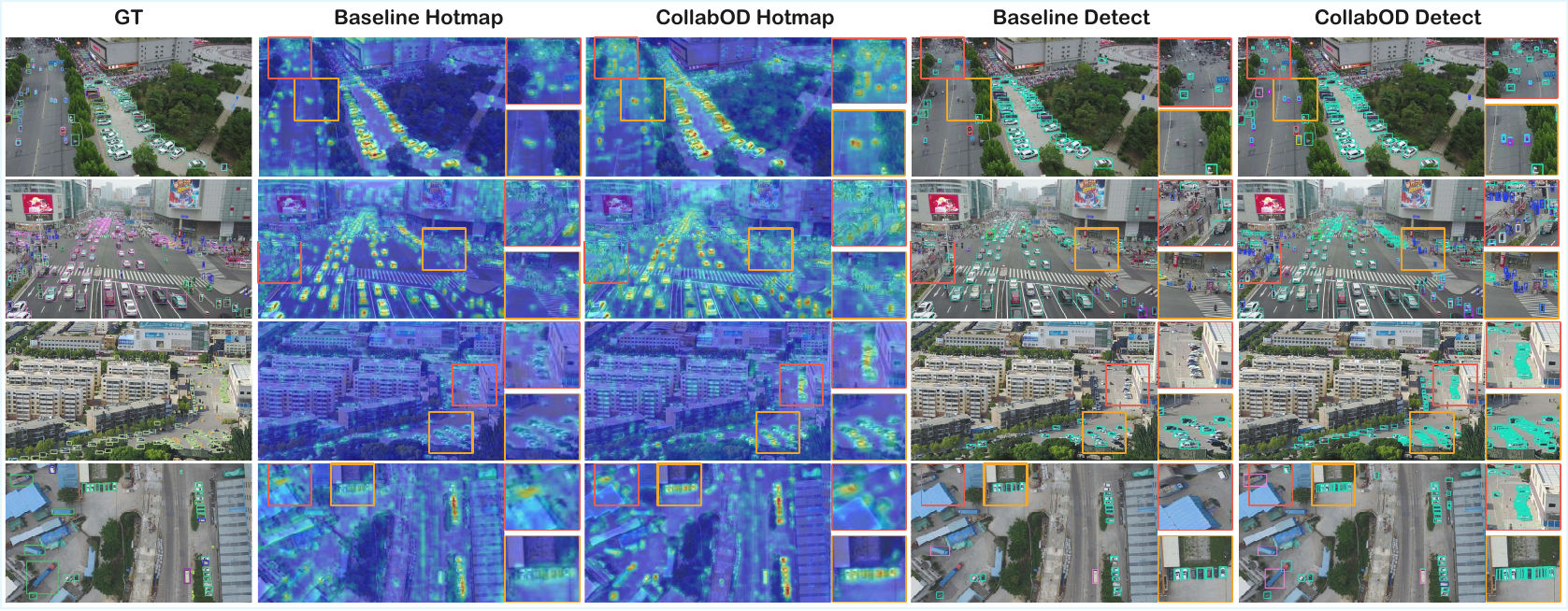}
  \caption{Qualitative comparison between the baseline and CollabOD on VisDrone-2019-DET. In complex aerial scenes with predominantly small objects, CollabOD shows fewer missed detections and more accurate localization than the baseline model.}
  \label{fig:vis}
\end{figure*}

\begin{table}[t]

\caption{
Comparative results on the \textbf{VisDrone-2019-DET}~\cite{vsdrone} dataset. Results for published methods are taken from their original papers when available, while YOLO-series baselines are evaluated under our controlled protocol. \textbf{Bold} and \underline{underline} denote the best and second-best results, respectively.}
\label{tab:visdrone_det}
\centering
\scriptsize
\setlength{\tabcolsep}{2pt}
\renewcommand{\arraystretch}{1.25}
\begin{tabularx}{\columnwidth}{%
X
S[table-format=2.1]
S[table-format=3.1]
S[table-format=2.1]
S[table-format=2.1]
S[table-format=2.1]
}
\specialrule{1.2pt}{0pt}{0pt}
Model & {Params} & {GFLOPs} & {AP$_{50} \uparrow$} & {AP$_{75} \uparrow$} & {AP$_{50:95} \uparrow$} \\
\specialrule{0.6pt}{0pt}{0pt}

\mbox{PP-YOLOE-SOD-L}~\cite{ppyoloe} & 42.2 & 120.5 & 48.5 & \multicolumn{1}{c}{\underline{30.4}} & 29.7 \\
CFPT~\cite{du2025cross}          & 56.3 & 297.6 & 38.0 & 23.1 & 22.6 \\
QueryDet~\cite{yang2022querydet}     & 33.9 & 212.0 & 48.1 & 28.8 & 28.3 \\
UAV-OD~\cite{wang2023generalized}        & \multicolumn{1}{c}{--} & \multicolumn{1}{c}{--} & 47.6 & 21.6 & 24.4 \\
UAV-DETR-R18~\cite{zhang2025uav}  & \multicolumn{1}{c}{\underline{20.0}} & 77.0  & 48.8 & 29.2 & 29.8 \\
UAV-DETR-R50~\cite{zhang2025uav}  & 42.0 & 170.0 & 51.1 & 30.4 & {\bfseries 31.5} \\
BRSTD-L~\cite{10700854}       & {\bfseries6.3} & 220.2 & {\bfseries58.0} & 19.5 & 26.1 \\
UAV-MaLO~\cite{wei2025uav}      & 21.6 & 73.6 & 49.9 & \multicolumn{1}{c}{--} & \multicolumn{1}{c}{\underline{30.1}} \\
YOLO11-M      & 20.1 & 68.2 & 43.3 & 24.7 & 26.3\\
YOLO11-M-P2 & 20.7 & 91.3 & 46.4 & 25.3 & 27.3 \\
YOLOv12-M~\cite{tian2025yolo12}     & 20.2 & \multicolumn{1}{c}{\underline{67.5}} & 33.6 & 18.1 & 19.2 \\
YOLOv12-M-P2~\cite{tian2025yolo12}  & \multicolumn{1}{c}{\underline{20.0}} & 77.7 & 36.2 & 24.2 & 21.0 \\
YOLO26-M~\cite{yolo26_ultralytics}      & 20.4 & 67.9  & 33.2 & 15.4 & 18.6 \\
YOLO26-M-P2~\cite{yolo26_ultralytics}   & 21.1 & 91.4  & 34.1 & 17.6 & 21.4 \\

\specialrule{0.6pt}{0pt}{0pt}
\rowcolor{blue!8}
{\bfseries CollabOD(Ours)} & 29.9 & {\bfseries \textcolor{red}{65.5}} & \multicolumn{1}{c}{\underline{52.4}} & {\bfseries30.8} & 29.9 \\

\specialrule{1.2pt}{0pt}{0pt}
\end{tabularx}
\end{table}

\subsection{Results on VisDrone Dataset}

\subsubsection{Comparative Results}
We compare CollabOD with recent and representative detectors on the VisDrone-2019-DET benchmark. The quantitative results are summarized in Table~\ref{tab:visdrone_det}.

With 20.9M parameters and 65.5 GFLOPs, CollabOD achieves 52.4 AP\textsubscript{50}, 30.8 AP\textsubscript{75}, and 29.9 AP\textsubscript{50:95}. Within this reported comparison, CollabOD attains the highest AP\textsubscript{75}, indicating improved localization stability under stricter IoU thresholds. This gain is consistent with explicitly strengthening localization-related structural cues and improving feature consistency prior to multi-scale fusion, while maintaining low computational cost.

Compared with the widely adopted YOLO11-M-P2, CollabOD improves AP\textsubscript{50} from 46.4 to 52.4 and AP\textsubscript{75} from 25.3 to 30.8, corresponding to gains of 6.0 and 5.5 percentage points, respectively. Meanwhile, the computational cost is reduced from 91.3 GFLOPs to 65.5 GFLOPs. This result indicates that enhancing localization-related structural information and calibrating heterogeneous feature streams prior to fusion can improve high-quality localization without increasing inference complexity.

In comparison with transformer-based approaches such as UAV-DETR-R50, which requires 170.0 GFLOPs, CollabOD achieves competitive detection performance with substantially lower computational overhead. These results suggest that the proposed framework provides strong localization capability while maintaining practical efficiency, which is particularly important for UAV-based deployment scenarios.

These results position CollabOD as an efficient and accurate solution for UAV-based small object detection.

To further validate its effectiveness, we present qualitative comparisons in Fig.~\ref{fig:vis}.
Even in cluttered scenes with densely distributed small objects, CollabOD maintains concentrated activation responses and stable localization, producing fewer missed detections and more precise bounding boxes.

\subsubsection{Reliability Check}
To keep the evaluation protocol clear, we use the VisDrone table only for the final model comparison and report the component-wise ablation on AI-TOD in Section~\ref{sec:aitod_ablation}. This separation keeps the VisDrone comparison focused on final-model metrics and prevents the ablation analysis from mixing different experiment settings.

\subsection{Results on UAVDT Dataset}

\begin{table}[t]
\caption{Comparison on the \textbf{UAVDT}~\cite{uavdt} dataset. 
The best and second-best results among the compared methods are highlighted in \textbf{bold} and \underline{underlined}, respectively.}
\label{tab:uavdt}
\centering
\renewcommand{\arraystretch}{1.2}

\begin{tabular}{
l
S[table-format=2.1]
S[table-format=3.2]
S[table-format=2.1]
}
\specialrule{1.2pt}{0pt}{0pt}
Model & {AP$_{50} \uparrow$} & {AP$_{75} \uparrow$} & {AP$_{50:95} \uparrow$} \\
\specialrule{0.6pt}{0pt}{0pt}

ClusDet~\cite{Yang_2019_ICCV} & 26.5 & 12.5 & 13.7 \\
GLSAN~\cite{9305976} & 28.1 & {\bfseries 18.8} & 17.0 \\
DREN~\cite{zhang2019fully} & \multicolumn{1}{c}{--} & \multicolumn{1}{c}{--} & 15.1 \\
GFL~\cite{li2020generalized} & 29.5 & \multicolumn{1}{c}{\underline{17.9}} & 16.9 \\
CEASC~\cite{du2023adaptive} & \multicolumn{1}{c}{\underline{30.9}} & 17.8 & \multicolumn{1}{c}{\underline{17.1}} \\

\specialrule{0.6pt}{0pt}{0pt}
\rowcolor{blue!8}
{\bfseries CollabOD(Ours)} & {\bfseries 31.2} & \multicolumn{1}{c}{\underline{17.9}} & {\bfseries 17.4} \\
\specialrule{1.2pt}{0pt}{0pt}
\end{tabular}
\end{table}

\begin{figure}[t]
  \centering
  \includegraphics[
    width=\linewidth]{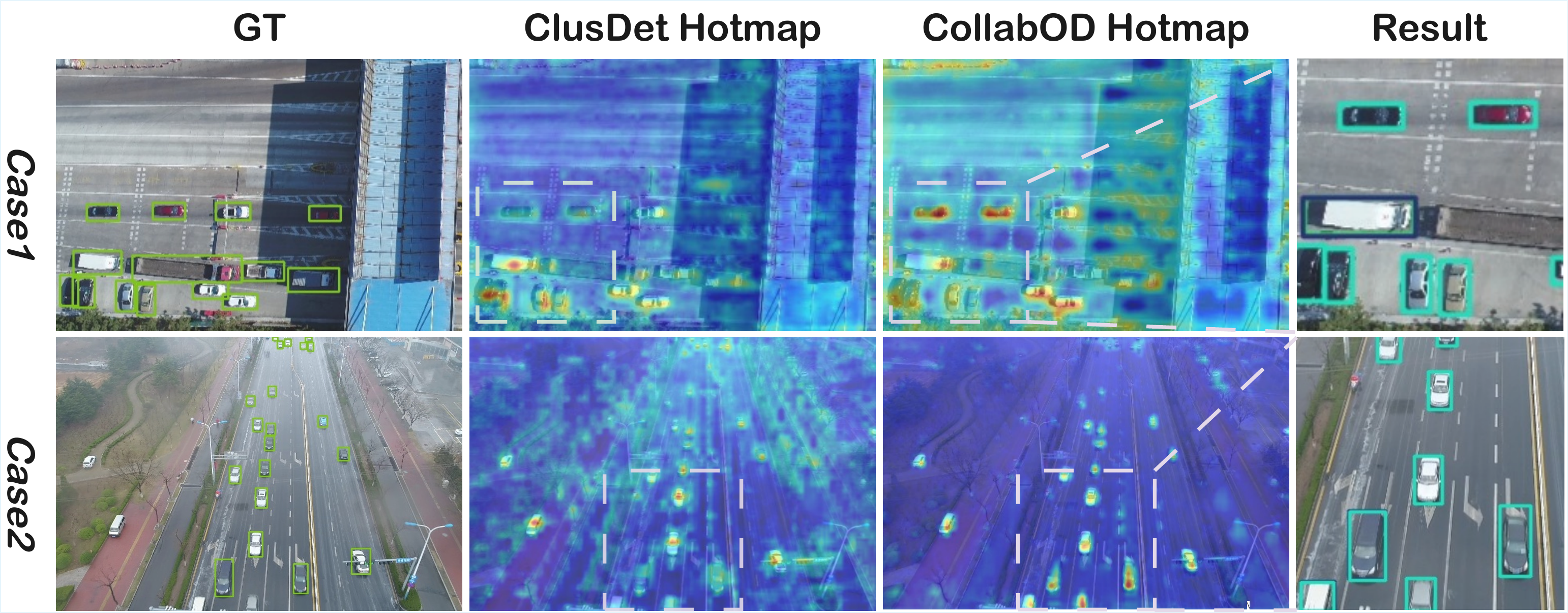}
  \caption{CollabOD localizes crowded UAVDT vehicles more consistently than ClusDet. Two representative cases are selected, and a focused comparison is conducted on the highlighted regions.}
  \label{fig:uavdt}
\end{figure}

On the UAVDT benchmark, the comparative results are reported in Table~\ref{tab:uavdt}. CollabOD achieves 31.2 AP\textsubscript{50}, 17.9 AP\textsubscript{75}, and 17.4 AP\textsubscript{50:95}, obtaining the best AP\textsubscript{50} and AP\textsubscript{50:95} among the compared methods. In addition, it ranks second on AP\textsubscript{75}, indicating stable localization performance under stricter IoU thresholds.

These results demonstrate that the proposed framework generalizes effectively to traffic-oriented UAV scenarios beyond the primary benchmark. To further examine the model behavior on the running traffic case, Fig.~\ref{fig:uavdt} compares the responses of ClusDet and CollabOD using the same visualization protocol as in the VisDrone experiments. The visualizations show that CollabOD generates more focused responses around object regions and maintains clearer separation from surrounding background areas, which is consistent with the improved quantitative performance reported in Table~\ref{tab:uavdt}.

\subsection{Results on AI-TOD Dataset}

\subsubsection{Comparative Results}
On the AI-TOD benchmark, the comparative results are summarized in Table~\ref{tab:aitod}. CollabOD achieves 45.4 AP$_{50}$ and 20.0 AP$_{50:95}$, obtaining the best performance among all YOLO-series models on both metrics. Compared with the strongest baseline, it improves AP$_{50}$ by 0.7 points over YOLOv12-M-P2 and is 0.5 points above the second-best AP$_{50:95}$ value of 19.5 achieved by YOLO11-M-P2.

In terms of efficiency, although CollabOD introduces slightly more parameters at 29.9M, it achieves the lowest computational cost of 65.5 GFLOPs and the highest inference speed of 137 FPS among the evaluated YOLO-series models, demonstrating a favorable accuracy--efficiency trade-off.

These results indicate that the proposed collaborative detection framework effectively enhances small-object detection performance on AI-TOD while maintaining competitive real-time capability.

\begin{table}[t]
\centering
\footnotesize
\setlength{\tabcolsep}{3.5pt}
\renewcommand{\arraystretch}{1.1}

\caption{Comparison on the \textbf{AI-TOD}~\cite{wang2021tiny} dataset. 
The best results among YOLO-series models are highlighted in \textbf{bold}.}
\label{tab:aitod}

\begin{tabular}{
l
S[table-format=2.1]
S[table-format=2.1]
S[table-format=2.1]
S[table-format=3.1]
S[table-format=3.0]
}
\toprule
Model 
& {AP$_{50}$} 
& {AP$_{50:95}$} 
& {Params} 
& {GFLOPs} 
& {FPS} \\
\midrule

YOLOv8-M-P2~\cite{yolov8_ultralytics}  
& 44.1 & 19.3 & 25.0 & 99.0 & 92 \\

YOLOv10-M-P2~\cite{THU-MIGyolov10}  
& 43.9 & 18.7 & 23.2 & 142.5 & 127 \\

YOLO11-M-P2~\cite{yolo11_ultralytics}  
& 44.5 & 19.5 & 20.7 & 91.3 & 101 \\

YOLOv12-M-P2~\cite{tian2025yolo12}  
& 44.7 & 17.2 & \textbf{20.0} & 94.4 & 107 \\

\midrule
\rowcolor{blue!8}
\textbf{CollabOD} 
& \textbf{45.4} 
& \textbf{20.0} 
& 29.9 
& \textbf{65.5} 
& \textbf{137} \\
\bottomrule
\end{tabular}
\end{table}

\subsubsection{Ablation Studies}
\label{sec:aitod_ablation}

On the AI-TOD dataset, we conduct an incremental ablation study to validate the progressive design of the proposed framework, as shown in Table~\ref{tab:aitod_ablation}. Starting from the baseline model, which achieves 44.5 AP$_{50}$ and 19.5 AP$_{50:95}$ with 91.3 GFLOPs and 101 FPS, we progressively incorporate DPF-Stem, DABlock, BRM, and the UDA Head.

\begin{table}[t]
\caption{Incremental ablation study on the \textbf{AI-TOD}~\cite{wang2021tiny} dataset. Components are progressively added to validate the final design. Bold values indicate the best performance.}
\label{tab:aitod_ablation}
\centering
\scriptsize
\setlength{\tabcolsep}{1.5pt}
\renewcommand{\arraystretch}{1.05}

\begin{tabular}{
cccc
S[table-format=2.1]
S[table-format=2.1]
S[table-format=3.1]
S[table-format=3.0]
}
\toprule

\multirow{2}{*}{DPF}
& \multirow{2}{*}{DAB}
& \multirow{2}{*}{BRM}
& \multirow{2}{*}{UDA}
& \multicolumn{2}{c}{Detection}
& \multicolumn{2}{c}{Complexity} \\

\cmidrule(lr){5-6}
\cmidrule(lr){7-8}

& & & 
& {AP$_{50}$} 
& {AP$_{50:95}$} 
& {GFLOPs} 
& {FPS}\\

\midrule

 &  &  &  & 44.5 & 19.5 & 91.3 & 101 \\
\cmark &  &  & & 43.2 & 18.4 & 51.2 & 110 \\
\cmark & \cmark &  &  & 43.7 & 18.8 & 68.2 & 107 \\
\cmark & \cmark & \cmark & & 44.2 & 19.3 & 74.8 & 118 \\
\rowcolor{blue!8}
\cmark & \cmark & \cmark & \cmark 
& \bfseries 45.4 
& \bfseries 20.0 
& \bfseries 65.5 
& \bfseries 137 \\

\bottomrule
\end{tabular}
\end{table}
Introducing DPF-Stem reduces the computational cost from 91.3 GFLOPs to 51.2 GFLOPs and improves the inference speed from 101 FPS to 110 FPS, but AP$_{50}$ decreases from 44.5 to 43.2. This result indicates that early structural preservation alone improves efficiency but is not sufficient to maintain detection accuracy. Adding DABlock and BRM gradually recovers AP$_{50}$ and AP$_{50:95}$, suggesting that dense aggregation and bilateral calibration compensate for representation loss introduced by the lightweight stem.

Finally, integrating the UDA Head yields the full CollabOD model, achieving 45.4 AP$_{50}$ and 20.0 AP$_{50:95}$ while reducing the computational cost to 65.5 GFLOPs and increasing the inference speed to 137 FPS. Since this table follows an incremental design, it supports the combined effect of the full design and the progressive recovery from the lightweight stem.

\subsubsection{Discussion}
The results across the three benchmarks indicate that CollabOD is effective when small objects appear with dense layouts, weak texture, and large scale variation. The gains at stricter IoU thresholds on VisDrone suggest improved box alignment, while the UAVDT visual comparison shows more stable responses for small vehicles in crowded traffic scenes. The AI-TOD ablation further shows that dense aggregation, bilateral recalibration, and uncertainty-aware prediction recover localization quality after the lightweight stem reduces early computation.

% \section{CONCLUSIONS}
% We have presented CollabOD, a lightweight framework for UAV small object detection that improves localization stability by preserving structural cues and calibrating feature streams prior to fusion. Experiments on VisDrone~\cite{vsdrone}, UAVDT~\cite{uavdt}, and AI-TOD~\cite{wang2021tiny} demonstrate improved performance under stricter IoU thresholds with competitive efficiency, showing a practical trade-off among accuracy, localization quality, and inference speed. Future work will explore real-time onboard deployment under hardware constraints and integration with downstream aerial tasks such as multi-object tracking and collaborative UAV perception. 

\section{CONCLUSION}
We have presented CollabOD, a lightweight collaborative framework for UAV small object detection that improves localization stability by preserving structural cues and calibrating heterogeneous feature streams before fusion. By combining the DPF-Stem, DABlock, BRM, and UDA Head, the proposed method enhances boundary- and detail-aware representations while maintaining efficient inference through lightweight design and re-parameterization. This design enables the detector to better handle weak textures, dense object layouts, and large-scale variations commonly observed in UAV imagery.

Experiments on VisDrone, UAVDT, and AI-TOD demonstrate improved performance under stricter IoU thresholds with competitive efficiency, showing a practical trade-off among accuracy, localization quality, and inference speed. The quantitative results and qualitative visualizations further verify that CollabOD produces more stable responses and more accurate localization for small objects in challenging aerial scenes. Future work will explore real-time onboard deployment under hardware constraints and integration with downstream aerial tasks such as multi-object tracking and collaborative UAV perception.

% \section*{ACKNOWLEDGMENT}

{
    \small
    \bibliographystyle{ieeenat_fullname}
    \bibliography{references}
}

\clearpage
\appendix
\twocolumn[
\begin{@twocolumnfalse}
\begin{center}
    {\LARGE \textbf{Technical Appendices and Supplementary Material}}
\end{center}
\vspace{1em}
\end{@twocolumnfalse}
]
\section{Supplementary Architectural Details}
\label{sec:supp_arch_details}

This section clarifies how the main modules of CollabOD are connected in the detector. We focus on where each module is inserted, which feature stream it acts on, and which type of small-object degradation it addresses.

\subsection{Overall Placement and Stage-wise Roles}
\label{sec:overall_placement}

CollabOD uses the YOLO11-M-P2 detector as the base framework and adds the proposed modules at different stages. The insertion points are chosen according to where small-object information is most likely to be weakened. Table~\ref{tab:module_placement} summarizes the role of each module.

DPF-Stem is placed at the input stage, before repeated downsampling removes weak local responses. DABlock is used inside the backbone to keep structural cues available during deeper feature propagation. BRM operates before cross-path fusion, where two feature streams may have different response magnitudes or spatial preferences. UDA Head takes the calibrated multi-scale features and predicts boxes with a localization-oriented head.

The order of these modules follows the degradation process of small UAV objects: weak details are first preserved, then propagated through the backbone, then calibrated before fusion, and finally used by the prediction head for localization. In this sense, CollabOD is not a direct stack of unrelated blocks. Each module targets a different stage of the feature pipeline, and the final model depends on their ordered cooperation.

\begin{table}[!th]
\caption{Functional placement of the proposed modules in CollabOD.}
\label{tab:module_placement}
\centering
\scriptsize
\setlength{\tabcolsep}{2.6pt}
\renewcommand{\arraystretch}{1.08}
\resizebox{\columnwidth}{!}{%
\begin{tabular}{llll}
\toprule
Module & Stage & Main Role & Target Issue \\
\midrule
DPF-Stem & Early stem & Detail-preserving encoding & Early structural loss \\
DABlock & Backbone & Dense local aggregation & Detail attenuation \\
BRM & Pre-fusion & Bilateral path calibration & Cross-path inconsistency \\
UDA Head & Detection head & Detail-aware prediction & Localization instability \\
\bottomrule
\end{tabular}%
}
\end{table}

\subsection{Structural Detail Preservation}
\label{sec:structural_preservation}

Small objects in UAV images often contain only weak contours, corners, and local texture gradients. Once these cues are suppressed by early downsampling, later fusion layers have little information to recover. We therefore use DPF-Stem at the input stage to reduce the loss of such early structural responses.

DPF-Stem separates early features into complementary pathways before fusion. These pathways are not meant to form independent semantic branches. Their purpose is more local: before the first strong feature compression, they retain different early responses so that weak boundaries and gradients are less likely to be discarded. The fused output then provides a richer low-level representation for the following backbone.

DABlock plays a different role. Even when early details are preserved, they can still fade during hierarchical propagation. DABlock reuses intermediate responses through dense local aggregation and lightweight refinement. This keeps shallow structural cues involved in deeper feature construction, which is important for small-object localization.

Thus, DPF-Stem and DABlock address different parts of the same problem. DPF-Stem protects early evidence, while DABlock helps carry that evidence forward. This separation makes the backbone less dependent on a single early feature map to preserve all small-object cues.

\subsection{Cross-Path Feature Calibration}
\label{sec:cross_path_calibration}

In a collaborative multi-path detector, different paths can have different receptive fields, activation distributions, and semantic biases. Direct addition or concatenation treats these streams as already compatible. This assumption is often weak in dense UAV scenes, where a small spatial mismatch or an imbalanced response magnitude can affect box regression.

BRM calibrates the two streams before they are fused. It first aligns them into a common channel space and then estimates bilateral spatial weights from their joint representation. Each path is reweighted with information from both paths, rather than being modulated independently. Learnable channel-wise scaling further adjusts the response magnitude of each stream before fusion.

This makes BRM different from a standard single-stream attention block: it calibrates two feature paths before fusion. The spatial masks support location-specific path selection, while the channel scalars reduce the chance that one path dominates simply because its activation magnitude is larger. This is useful for small objects, whose weak structural responses can easily be overwhelmed by background activations or high-level semantic responses.

\subsection{Localization-Aware Prediction Head}
\label{sec:uda_head_detail}

UDA Head uses the preserved and calibrated features for final prediction. Each input feature is first projected into a shared hidden space, and a shared detail-enhancement block is applied across detection scales. This encourages the regression branch to use boundary-sensitive responses in a consistent way across different feature levels.

The head keeps the box regression and classification branches decoupled. For regression, distribution-based box prediction improves localization under small-object scale ambiguity. The shared detail-enhancement block avoids building a heavy independent prediction branch for each scale, so the head improves localization sensitivity while keeping the inference cost controlled.

Overall, CollabOD follows a stage-aware feature pipeline. It preserves weak structural evidence early, keeps it active through backbone propagation, calibrates heterogeneous streams before fusion, and finally uses the refined features for localization-aware prediction.

%%%%%%%%%%%%%%%%%%%%%%%%%%%%%%%%%%%%%%%%%%%%%%%%%%%%%%%%%%%%%%%%%%%%%%%%%%%%%%%%
\section{Supplementary Experimental Analysis}
\label{sec:supplementary_experiments}

This section reports three quantitative analyses and two additional qualitative visualizations. The quantitative part covers training stability, replacement of the cross-path calibration module, and independent component ablation. The qualitative part adds visual evidence on VisDrone and CARPK~\cite{hsieh2017drone} to show how the learned responses align with small-object regions in crowded aerial scenes.

%%%%%%%%%%%%%%%%%%%%%%%%%%%%%%%%%%%%%%%%%%%%%%%%%%%%%%%%%%%%%%%%%%%%%%%%%%%%%%%%
\subsection{Training Stability Across Random Seeds}
\label{sec:seed_stability}

We train representative models with three random seeds and report the mean and standard deviation in Table~\ref{tab:seed_stability}. All runs use the same input resolution, optimizer, training schedule, batch size, and evaluation code as the main experiments. The compared models include the YOLO11-M-P2 baseline, two ablated variants, and the full CollabOD model.

Table~\ref{tab:seed_stability} shows that CollabOD remains better than the YOLO11-M-P2 baseline across repeated training. The standard deviations are small on AP$_{50}$, AP$_{50:95}$, and FPS, which suggests that the improvement is not highly sensitive to random initialization. The two ablated variants also stay below the full model, consistent with the component analysis in the main paper.

\begin{table}[H]
\caption{Training stability analysis on AI-TOD over three random seeds. We report mean and standard deviation.}
\label{tab:seed_stability}
\centering
\scriptsize
\setlength{\tabcolsep}{3.5pt}
\renewcommand{\arraystretch}{1.1}
\resizebox{\columnwidth}{!}{%
\begin{tabular}{lccc}
\toprule
Model & AP$_{50}$ & AP$_{50:95}$ & FPS \\
\midrule
YOLO11-M-P2 & 44.5 $\pm$ 0.2 & 19.5 $\pm$ 0.1 & 101 $\pm$ 2 \\
CollabOD w/o BRM & 44.1 $\pm$ 0.1 & 19.2 $\pm$ 0.3 & 120 $\pm$ 1 \\
CollabOD w/o UDA Head & 44.2 $\pm$ 0.3 & 19.3 $\pm$ 0.2 & 118 $\pm$ 3 \\
\rowcolor{blue!8}
\textbf{CollabOD} & \textbf{45.4 $\pm$ 0.1} & \textbf{20.0 $\pm$ 0.3} & \textbf{137 $\pm$ 3} \\
\bottomrule
\end{tabular}%
}
\end{table}

%%%%%%%%%%%%%%%%%%%%%%%%%%%%%%%%%%%%%%%%%%%%%%%%%%%%%%%%%%%%%%%%%%%%%%%%%%%%%%%%
\subsection{Comparison with Alternative Cross-Path Calibration Strategies}
\label{sec:brm_replacement}

We further replace BRM with several common fusion or attention choices while keeping the rest of the architecture unchanged. The alternatives include direct addition, concatenation followed by convolution, channel-only gating, and generic channel-spatial gating.

Table~\ref{tab:brm_replacement} shows the expected accuracy-efficiency trade-off. Direct addition is fast, but it does not handle feature mismatch before fusion. Concatenation followed by convolution improves the fused representation but increases computation. Channel-only gating provides feature selection, while channel-spatial gating adds spatial modulation. However, neither explicitly estimates bilateral masks from the joint two-path representation. BRM gives the best AP$_{50}$ and AP$_{50:95}$ while keeping GFLOPs and FPS competitive, which supports its role as a pre-fusion path calibration module.

\begin{table}[H]
\caption{Comparison of different cross-path calibration strategies on AI-TOD. All variants use the same overall architecture except for the cross-path calibration module.}
\label{tab:brm_replacement}
\centering
\scriptsize
\setlength{\tabcolsep}{3.5pt}
\renewcommand{\arraystretch}{1.1}
\resizebox{\columnwidth}{!}{%
\begin{tabular}{lcccc}
\toprule
Strategy & AP$_{50}$ & AP$_{50:95}$ & GFLOPs & FPS \\
\midrule
Direct Add & 44.5 & 19.4 & 64.8 & 142 \\
Concat + Conv & 44.8 & 19.6 & 68.1 & 132 \\
Channel Gate & 44.9 & 19.6 & 66.2 & 135 \\
Channel-Spatial Gate & 45.1 & 19.8 & 68.6 & 129 \\
\rowcolor{blue!8}
\textbf{BRM} & \textbf{45.4} & \textbf{20.0} & 65.5 & 137 \\
\bottomrule
\end{tabular}%
}
\end{table}

%%%%%%%%%%%%%%%%%%%%%%%%%%%%%%%%%%%%%%%%%%%%%%%%%%%%%%%%%%%%%%%%%%%%%%%%%%%%%%%%
\subsection{Independent and Leave-One-Out Component Ablation}
\label{sec:independent_ablation}

The main paper reports an incremental ablation study. That setting is useful for showing how the final model is progressively built, but it does not fully separate the contribution of each module. We therefore include independent and leave-one-out ablations in Table~\ref{tab:independent_ablation}.

The ``only'' variants test whether a single module is sufficient to improve the baseline. The ``w/o'' variants test whether each module remains necessary when the other components are present. This gives a clearer view of both standalone contribution and full-model dependency.

DPF-Stem alone improves efficiency but reduces accuracy, which indicates that early lightweight encoding needs the later modules to recover full detection quality. DABlock, BRM, and UDA Head each bring moderate gains when used independently. In the leave-one-out setting, removing any major component reduces the final performance. The full model therefore benefits from the interaction among early detail preservation, structural propagation, cross-path calibration, and localization-aware prediction.

\begin{figure*}[p]
\centering
\includegraphics[width=0.93\textwidth]{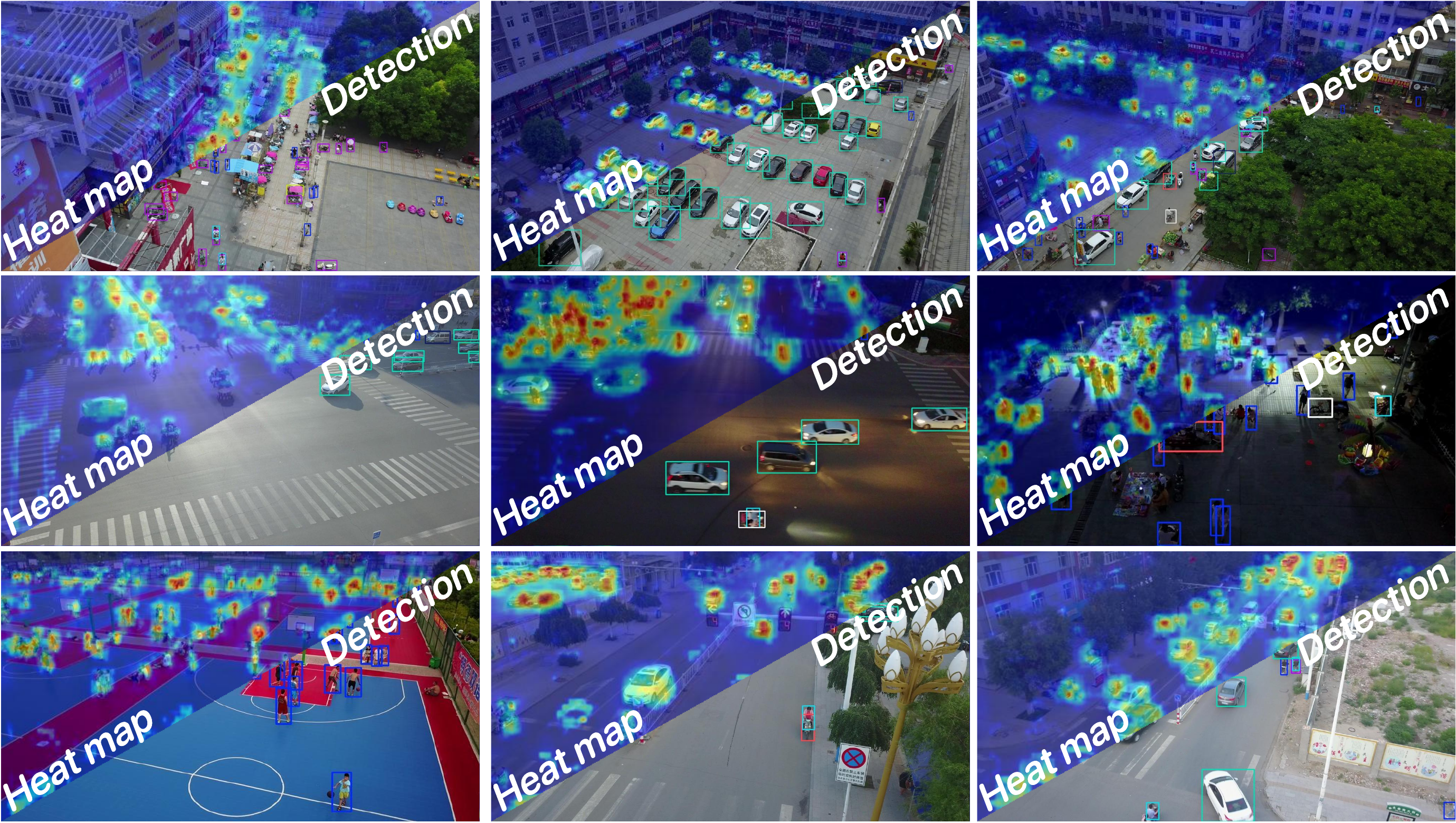}
\vspace{0.8em}
\includegraphics[width=0.93\textwidth]{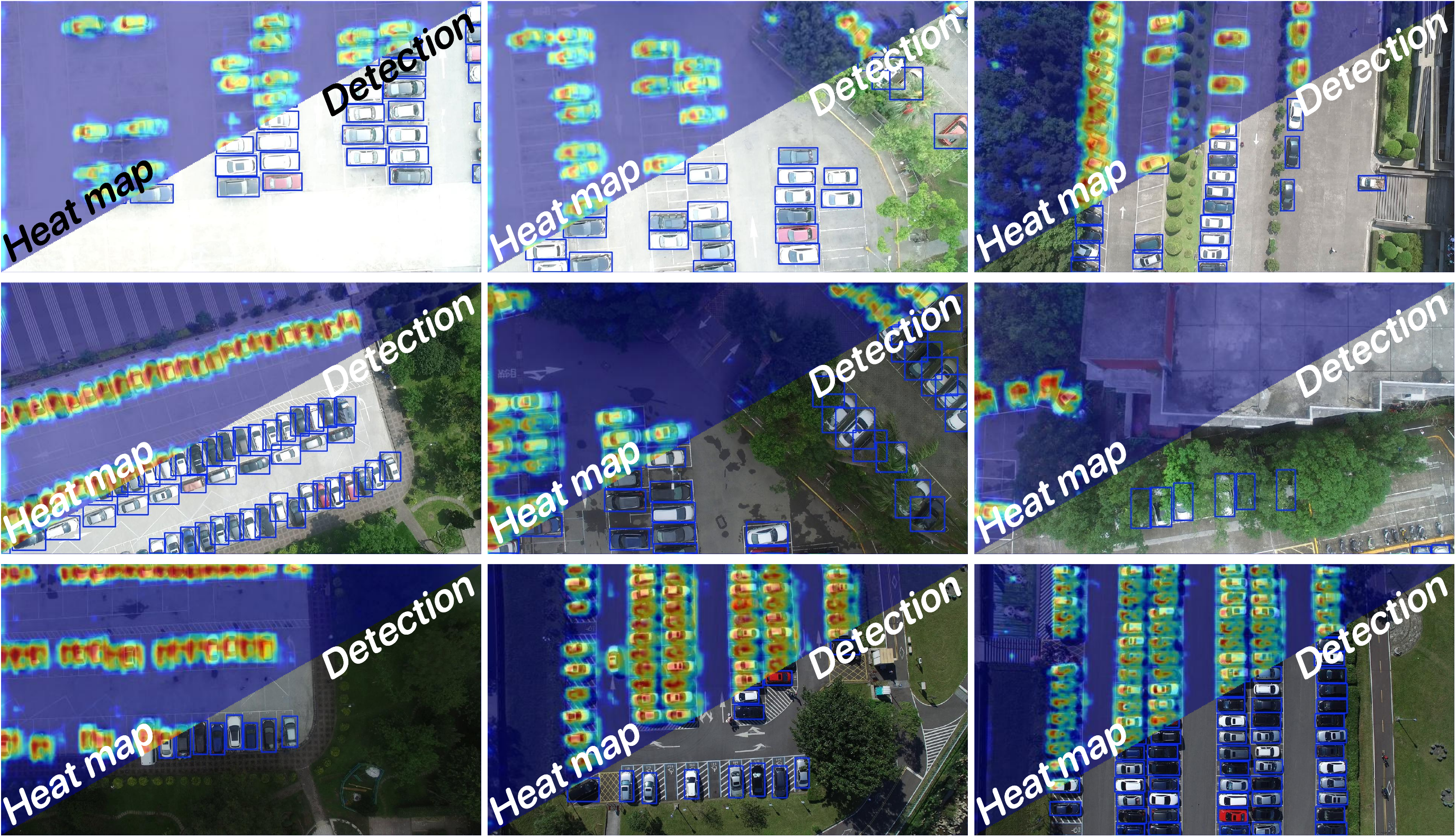}
\caption{Additional qualitative visualizations on VisDrone (top) and CARPK~\cite{hsieh2017drone} (bottom). The diagonal split shows the heat-map response on one side and the detection result on the other side. VisDrone covers crowded UAV scenes with scale variation and cluttered backgrounds, while CARPK focuses on high-density vehicle layouts.}
\label{fig:qual_visualization}
\end{figure*}

\begin{table}[H]
\caption{Single-column independent and leave-one-out ablation studies on AI-TOD.}
\label{tab:independent_ablation}
\centering
\scriptsize
\setlength{\tabcolsep}{2.8pt}
\renewcommand{\arraystretch}{1.08}
\resizebox{\columnwidth}{!}{%
\begin{tabular}{lcccccc}
\toprule
Setting & DPF & DAB & BRM & UDA & AP$_{50}$ & AP$_{50:95}$ \\
\midrule
Baseline &  &  &  &  & 44.5 & 19.5 \\
DPF only & \cmark &  &  &  & 43.2 & 18.4 \\
DAB only &  & \cmark &  &  & 44.7 & 19.6 \\
BRM only &  &  & \cmark &  & 44.8 & 19.7 \\
UDA only &  &  &  & \cmark & 45.0 & 19.8 \\
\midrule
Full w/o DPF &  & \cmark & \cmark & \cmark & 44.9 & 19.7 \\
Full w/o DAB & \cmark &  & \cmark & \cmark & 45.0 & 19.7 \\
Full w/o BRM & \cmark & \cmark &  & \cmark & 44.3 & 19.2 \\
Full w/o UDA & \cmark & \cmark & \cmark &  & 44.2 & 19.3 \\
\rowcolor{blue!8}
\textbf{Full CollabOD} & \cmark & \cmark & \cmark & \cmark & \textbf{45.4} & \textbf{20.0} \\
\bottomrule
\end{tabular}%
}
\end{table}

%%%%%%%%%%%%%%%%%%%%%%%%%%%%%%%%%%%%%%%%%%%%%%%%%%%%%%%%%%%%%%%%%%%%%%%%%%%%%%%%

\subsection{Additional Qualitative Visualization}
\label{sec:qual_visualization}

We also provide qualitative visualizations on VisDrone and CARPK~\cite{hsieh2017drone} in Fig.~\ref{fig:qual_visualization}. Each example is shown with a diagonal split: the heat-map response is shown on one side, and the detection result is shown on the other side. The goal is to complement the quantitative tables by checking whether the response distribution is spatially consistent with small-object locations in dense aerial scenes.

The VisDrone examples include more diverse object categories and background conditions, while CARPK mainly focuses on dense vehicle layouts. In both cases, high-response regions generally follow the small-object distribution. This observation is consistent with the intended behavior of CollabOD: preserving weak structural details and calibrating feature streams before prediction.

\subsection{Discussion}
\label{sec:supplementary_discussion}

These supplementary results mainly support four points. First, the seed-level results indicate that CollabOD is not highly sensitive to random initialization. Second, the replacement study shows that BRM is more suitable for this architecture than direct addition, concatenation-based fusion, channel-only gating, or generic channel-spatial gating. Third, the independent and leave-one-out ablations show that the full model depends on the cooperation among all proposed components, rather than on a single isolated modification. Fourth, the VisDrone and CARPK~\cite{hsieh2017drone} visualizations show that the response maps are spatially consistent with dense small-object regions, which supports the qualitative side of the proposed detail-preserving and calibration-based design.

Overall, the appendix supports the same design principle as the main paper: small-object detection in UAV images benefits from preserving weak structural details, keeping them active through backbone propagation, calibrating heterogeneous streams before fusion, and using a lightweight localization-aware prediction head. The added visualizations further show how this design behaves in both general UAV detection scenes and dense parking-lot scenarios.

\end{document}